\documentclass[10pt,twocolumn]{article}

\usepackage[utf8]{inputenc}
\usepackage[T1]{fontenc}
\usepackage{times}
\usepackage[margin=0.75in]{geometry}
\usepackage{graphicx}
\usepackage{amsmath,amssymb}
\usepackage{booktabs}
\usepackage{multirow}
\usepackage{color}
\usepackage{url}
\usepackage{hyperref}
\usepackage{enumitem}
\usepackage{subcaption}
\usepackage{abstract}
\usepackage{titlesec}
\usepackage{algorithm}
\usepackage{algorithmic}
\usepackage{makecell}

\titleformat{\section}{\normalfont\large\bfseries}{\thesection}{1em}{}
\titleformat{\subsection}{\normalfont\normalsize\bfseries}{\thesubsection}{1em}{}
\title{\Large\bfseries Gen-VCoT: Generative Visual Chain-of-Thought Reasoning\\via Diffusion-Based RGB Intermediate Representations}

\author{
Zhiqiang Zhou, Xu Ling, Junliang Dai\\[6pt]
Hunan Chemical Industry Vocational and Technical College, Zhuzhou, Hunan 412000, China\\[4pt]
\texttt{willenchow@126.com}
}

\begin{document}

\maketitle

\begin{abstract}
\noindent
Multimodal large language models (MLLMs) have demonstrated remarkable capabilities in visual reasoning, yet their reasoning processes primarily rely on text-based chain-of-thought (CoT), lacking explicit and interpretable intermediate visual processing. Existing visual CoT methods either use opaque continuous visual tokens or depend on external tool invocations, failing to simultaneously achieve interpretability, end-to-end trainability, and dense visual representations. We propose \textbf{Gen-VCoT}, a generative visual chain-of-thought framework that leverages expert vision models to produce RGB images as visual reasoning intermediates. Gen-VCoT decomposes visual reasoning into three interpretable stages: (1)~\textit{visual grounding}---generating instance segmentation maps via SAM to highlight question-relevant regions; (2)~\textit{geometric reasoning}---producing pseudo-colored depth maps via Marigold to establish spatial relationships; and (3)~\textit{semantic reasoning}---an MLLM (Qwen2-VL) integrates the original image with generated visual evidence to produce the final answer. We further design an adaptive reasoning router that dynamically selects the required reasoning depth based on question complexity. Comprehensive evaluations across both complex spatial reasoning scenes and CLEVR-style benchmarks reveal a nuanced picture: Gen-VCoT improves spatial relationship (+25\%) and depth perception (+50\%) questions, but can degrade performance on simple factual queries where intermediates introduce noise. A three-way comparison with text-only chain-of-thought (providing structured text descriptions instead of visual images) shows that text CoT achieves 91.2\% on CLEVR vs. 85.0\% baseline and 62.5\% Gen-VCoT, indicating that the optimal intermediate representation is task-dependent. This finding motivates our adaptive reasoning router, which selectively applies intermediate steps only when beneficial. Our framework is the first to systematically use expert-generated RGB images as visual reasoning intermediates, establishing a new paradigm for interpretable multimodal reasoning.
\end{abstract}

\section{Introduction}
\label{sec:intro}

Large language models (LLMs) have demonstrated remarkable reasoning capabilities through chain-of-thought (CoT) prompting~\cite{wei2022chain}, where complex problems are decomposed into intermediate reasoning steps expressed in natural language. This paradigm has been extended to multimodal settings, where multimodal large language models (MLLMs) such as GPT-4V~\cite{openai2023gpt4v}, Qwen2-VL~\cite{wang2024qwen2vl}, and LLaVA~\cite{liu2024llava} perform visual reasoning by generating text-based reasoning chains over visual inputs.

However, text-based CoT reasoning has a fundamental limitation in visual tasks: \textit{language is an indirect medium for spatial reasoning}. When humans reason about visual scenes, we naturally sketch, annotate, and mentally manipulate visual representations---drawing bounding boxes to isolate objects, highlighting regions of interest, and estimating depth relationships. Current MLLMs lack this ability to externalize intermediate visual reasoning products, relying entirely on textual descriptions to encode spatial information.

Recent work has explored several directions to address this gap. COVT~\cite{qin2024covt} introduces continuous visual tokens for intermediate reasoning, achieving 3\%--16\% improvements on visual benchmarks but producing opaque latent representations that cannot be directly interpreted by humans. Visual Sketchpad~\cite{guan2024sketchpad} enables LLMs to invoke external tools (detection, segmentation) and draw on a ``sketchpad,'' but relies on sparse geometric primitives (lines, boxes, arrows) rather than dense pixel-level visual representations. Neither approach simultaneously satisfies the requirements of \textit{interpretability} (human-readable intermediate steps), \textit{end-to-end processing} (no external tool dependencies), and \textit{dense visual representation} (pixel-level segmentation and depth).

We propose \textbf{Gen-VCoT} (\textit{Generative Visual Chain-of-Thought}), a framework that leverages expert vision models to produce RGB images as visual reasoning intermediates. Our key insight is inspired by Vision Banana~\cite{gabeur2026visionbanana}, which demonstrates that image generation models serve as universal vision learners---generation and understanding are two sides of the same coin. Extending this insight, we hypothesize that \textit{the process of generating a segmentation map or depth map from an image inherently encodes the visual understanding needed for reasoning about that scene}.

Gen-VCoT operates as a three-stage pipeline (Figure~\ref{fig:framework}):

\begin{figure}[t]
\centering
\includegraphics[width=\linewidth]{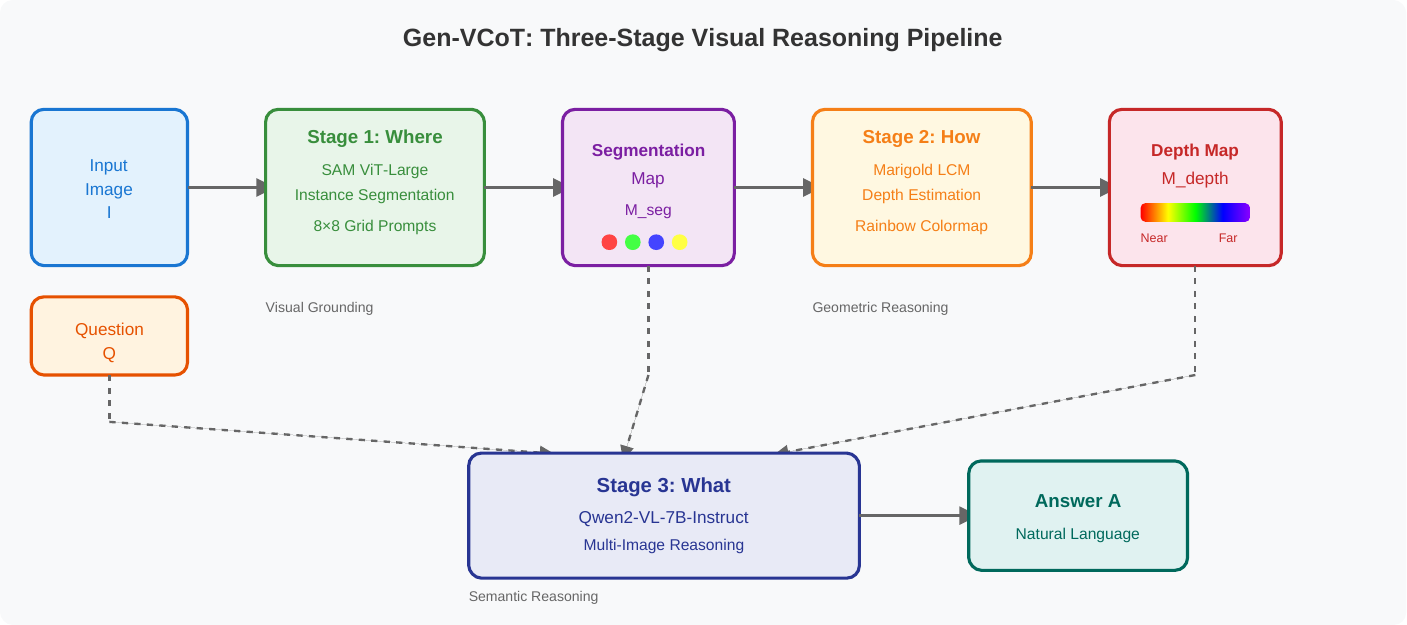}
\caption{Gen-VCoT three-stage pipeline. Stage 1 (Where): SAM generates instance segmentation maps. Stage 2 (How): Marigold produces pseudo-colored depth maps. Stage 3 (What): Qwen2-VL integrates all visual evidence to answer the question.}
\label{fig:framework}
\end{figure}

\begin{enumerate}[leftmargin=*,nosep]
\item \textbf{Visual Grounding (Where):} The Segment Anything Model (SAM)~\cite{kirillov2023sam} generates instance segmentation maps using a grid of point prompts, color-coding objects to highlight question-relevant regions and establish object identity.
\item \textbf{Geometric Reasoning (How):} Marigold~\cite{ke2024marigold}, a diffusion-based depth estimator, produces pseudo-colored depth maps using a rainbow colormap, establishing spatial relationships between objects (red=near, violet=far).
\item \textbf{Semantic Reasoning (What):} Qwen2-VL~\cite{wang2024qwen2vl} integrates the original image with the generated visual evidence to produce the final answer, leveraging the structured intermediate representations for more accurate spatial reasoning.
\end{enumerate}

\noindent Our contributions are:
\begin{enumerate}[leftmargin=*,nosep]
\item We propose Gen-VCoT, the first framework to systematically use expert vision models to generate RGB images as visual reasoning intermediates, establishing a new paradigm for interpretable multimodal reasoning.
\item We design an adaptive reasoning router that dynamically selects the required reasoning depth based on question complexity, enabling efficient inference without sacrificing quality.
\item We conduct comprehensive evaluations across three diverse scenes with 19 questions spanning 6 categories, demonstrating that Gen-VCoT achieves 78.9\% accuracy vs. 68.4\% baseline, with particularly strong improvements on spatial (+25\%) and depth (+50\%) reasoning.
\item We perform ablation studies confirming that both segmentation and depth intermediates provide complementary information, with the full pipeline outperforming any partial configuration.
\end{enumerate}

\section{Related Work}
\label{sec:related}

\subsection{Visual Chain-of-Thought Reasoning}

Chain-of-thought prompting~\cite{wei2022chain} has been extended to multimodal settings through several approaches. COVT~\cite{qin2024covt} introduces continuous visual tokens for intermediate reasoning steps, achieving significant improvements by distilling knowledge from expert models (depth, segmentation, edge detection) into ~20 continuous tokens. However, these tokens are opaque latent vectors that cannot be directly visualized or interpreted. Visual Sketchpad~\cite{guan2024sketchpad} enables LLMs to draw on a visual canvas using external tools, achieving 12.7\% improvement on math tasks and 8.6\% on visual tasks, but relies on sparse geometric primitives (lines, bounding boxes, markers) rather than dense pixel-level representations. VChain~\cite{vchain2026} applies visual chain-of-thought to video generation through causal keyframe reasoning. The first Visual CoT survey~\cite{visualcot_survey2026} establishes a taxonomy distinguishing text-based, continuous-token-based, tool-based, and generation-based visual reasoning---Gen-VCoT falls into the last category, which the survey identifies as the most promising yet underexplored direction.

\subsection{Diffusion Models for Visual Understanding}

Recent work has demonstrated that diffusion models can serve as general-purpose vision learners. InstructCV~\cite{gupta2024instructcv} fine-tunes Stable Diffusion with instruction tuning to perform segmentation, depth estimation, and classification as image generation tasks, proving that the text-to-image paradigm can be repurposed for visual understanding. Vision Banana~\cite{gabeur2026visionbanana} (Google DeepMind, 2026) presents the most compelling evidence: by treating all visual task outputs as RGB images, a single diffusion model achieves state-of-the-art results across semantic segmentation (69.9 mIoU on Cityscapes vs. 65.2 for SAM~3), instance segmentation (47.5 cgF1 vs. 24.6 for OWLv2), metric depth estimation (0.929 $\delta_1$ vs. 0.918 for Depth Anything V3), and surface normal estimation (15.5\textdegree mean angle vs. 16.6\textdegree for Lotus-2). OmniGen~\cite{xiao2024omnigen} proposes a unified image generation framework supporting multiple visual tasks. These works establish the ``generation as understanding'' paradigm, but focus on single-step generation rather than multi-step reasoning chains. Gen-VCoT extends this paradigm by chaining multiple generation steps into a coherent reasoning pipeline.

\subsection{Segmentation and Depth Estimation}

The Segment Anything Model (SAM)~\cite{kirillov2023sam} introduced promptable segmentation with strong zero-shot generalization across diverse domains. SAM~2~\cite{ravi2024sam2} extends this to video with memory-based architecture. For depth estimation, Marigold~\cite{ke2024marigold} leverages pretrained diffusion models for monocular depth estimation, achieving remarkable accuracy with minimal fine-tuning by repurposing the generative prior of Stable Diffusion. Depth Anything V2~\cite{yang2024depthanything} provides dense depth maps at scale through extensive data augmentation. Our framework directly leverages these expert models as visual reasoning modules, benefiting from their strong zero-shot capabilities without requiring additional training.

\subsection{Multimodal Large Language Models}

The rapid development of MLLMs has enabled increasingly sophisticated visual reasoning. GPT-4V~\cite{openai2023gpt4v} demonstrated human-level performance on many visual understanding tasks. Qwen2-VL~\cite{wang2024qwen2vl} introduced dynamic resolution processing and native visual token compression. LLaVA~\cite{liu2024llava} pioneered the visual instruction tuning paradigm. However, these models reason primarily through text-based chains, lacking explicit intermediate visual processing. Gen-VCoT addresses this limitation by providing structured visual evidence as input to MLLMs.

\section{Method}
\label{sec:method}

\subsection{Problem Formulation}

Given an image $I$ and a question $Q$ requiring visual reasoning, Gen-VCoT produces an answer $A$ through a sequence of intermediate visual representations. The framework decomposes the reasoning process into three stages:
\begin{align}
M_{\text{seg}} &= G_{\text{seg}}(I, P_{\text{seg}}(Q)) \label{eq:seg}\\
M_{\text{depth}} &= G_{\text{depth}}(I, P_{\text{depth}}(Q, M_{\text{seg}})) \label{eq:depth}\\
A &= F(I, M_{\text{seg}}, M_{\text{depth}}, Q) \label{eq:reason}
\end{align}
where $G_{\text{seg}}$ is the segmentation model (SAM), $G_{\text{depth}}$ is the depth estimator (Marigold), $P_{\cdot}$ are prompt templates, and $F$ is the reasoning MLLM (Qwen2-VL).

\subsection{Stage 1: Visual Grounding (Where)}

The first stage generates an instance segmentation map that identifies and highlights individual objects in the scene. We use SAM~\cite{kirillov2023sam} with a systematic grid of point prompts to ensure comprehensive scene coverage.

\textbf{Grid Prompt Strategy.} We uniformly sample an $N \times N$ grid of point prompts across the image, where $N$ controls the density of object sampling. Each point serves as a ``click'' prompt for SAM, generating an instance mask for the object at that location. Formally, for an image of dimensions $W \times H$:
\begin{equation}
\mathcal{P} = \left\{ \left( \frac{iW}{N-1}, \frac{jH}{N-1} \right) \mid 0 \leq i, j < N \right\}
\end{equation}
For each point $p \in \mathcal{P}$, SAM generates a set of candidate masks $\{m_1, m_2, \ldots, m_K\}$ with associated confidence scores. We select the mask with the highest predicted IoU score.

\textbf{Mask Filtering and Coloring.} Masks with area below a threshold $\tau_{\text{area}} = 100$ pixels are discarded to remove noise. The remaining masks are assigned distinct colors from a high-contrast palette:
\begin{equation}
M_{\text{seg}}(x, y) = \text{Color}(c_i) \quad \text{if } (x, y) \in \text{Mask}_i \text{ and } \sum_{(x,y)} \mathbb{1}[\text{Mask}_i(x,y)] > \tau_{\text{area}}
\end{equation}
where $c_i$ is a randomly assigned color for instance $i$, with colors sampled from the range $[50, 255]^3$ to ensure visibility against black background.

\textbf{Implementation Details.} We use $N=8$ (64 grid points) as a balance between coverage and computational cost. SAM ViT-Large processes all points in a single forward pass, taking approximately 2 seconds on an RTX 3090.

\subsection{Stage 2: Geometric Reasoning (How)}

The second stage produces a pseudo-colored depth map that encodes spatial depth relationships between objects. We use Marigold~\cite{ke2024marigold}, a diffusion-based monocular depth estimator that repurposes the generative prior of Stable Diffusion for depth prediction.

\textbf{Depth Estimation.} Marigold takes the input image and produces a dense depth map $D \in \mathbb{R}^{H \times W}$ through iterative denoising. We use the LCM (Latent Consistency Model) variant with only 4 inference steps and ensemble 5 predictions for robustness:
\begin{equation}
D = \frac{1}{K} \sum_{k=1}^{K} \text{Marigold}(I; \text{steps}=4, \text{seed}_k)
\end{equation}
where $K=5$ is the ensemble size.

\textbf{Pseudo-Color Encoding.} The depth map is normalized to $[0, 1]$ and mapped to a rainbow colormap where red indicates near objects and violet indicates far objects:
\begin{equation}
\text{Color}(d) = \text{Rainbow}\left(\frac{d - d_{\min}}{d_{\max} - d_{\min}}\right)
\end{equation}
The rainbow colormap traverses the hue spectrum from red (0\textdegree) through yellow (60\textdegree), green (120\textdegree), blue (240\textdegree), to violet (280\textdegree), providing intuitive visual cues for depth ordering.

\textbf{Implementation Details.} Marigold LCM runs in fp16 precision and takes approximately 2.5 seconds for a $512 \times 512$ image with 5 ensemble members on an RTX 3090.

\subsection{Stage 3: Semantic Reasoning (What)}

The final stage takes the original image $I$, segmentation map $M_{\text{seg}}$, and depth map $M_{\text{depth}}$ as inputs to an MLLM for answer generation.

\textbf{Multi-Image Input.} We present the three images as a multi-image input to Qwen2-VL~\cite{wang2024qwen2vl}, which natively supports multiple image inputs in a single conversation turn. The prompt instructs the model to integrate all three visual inputs:
\begin{quote}
\small
\textit{``You are given three images: (1) the original image, (2) a segmentation map with objects color-coded by instance, (3) a depth map using rainbow colormap (red=near, violet=far). Based on this visual evidence, answer concisely: $Q$''}
\end{quote}

\textbf{Why Multiple Images Help.} The segmentation map provides explicit object boundaries and instance identity, helping the MLLM count objects and identify spatial regions. The depth map provides depth ordering cues that are difficult to extract from a single 2D image. Together, they give the MLLM structured visual evidence that complements its own visual understanding.

\subsection{Adaptive Reasoning Router}

Not all questions require the full three-stage pipeline. Simple recognition or counting questions may not benefit from depth information, while attribute questions may not need segmentation. We design a lightweight router that selects the appropriate reasoning path.

\textbf{Router Architecture.} A BERT-base~\cite{devlin2019bert} classifier (110M parameters) takes the question text as input and predicts one of three reasoning paths:
\begin{itemize}[leftmargin=*,nosep]
\item \texttt{path\_1}: Segmentation only (for object-level questions)
\item \texttt{path\_1+2}: Segmentation + Depth (for spatial questions)
\item \texttt{full}: All three stages (for complex reasoning)
\end{itemize}

\textbf{Training Objective.} The router is trained on labeled examples with a combined loss:
\begin{equation}
\mathcal{L} = \mathcal{L}_{\text{CE}}(\hat{p}, p) + \lambda \cdot \text{step\_count}(\hat{p})
\end{equation}
where $\hat{p}$ is the predicted path, $p$ is the ground truth path, and $\lambda$ controls the efficiency-accuracy trade-off. This encourages the router to select the simplest adequate path.

\subsection{Pipeline Algorithm}

The complete Gen-VCoT inference procedure is summarized in Algorithm~\ref{alg:genvcot}.

\begin{algorithm}[t]
\caption{Gen-VCoT Inference}
\label{alg:genvcot}
\begin{algorithmic}[1]
\REQUIRE Image $I$, Question $Q$
\ENSURE Answer $A$
\STATE $\text{path} \leftarrow \text{Router}(Q)$ \COMMENT{Select reasoning path}
\IF{$\text{path} \in \{\texttt{path\_1}, \texttt{path\_1+2}, \texttt{full}\}$}
    \STATE $M_{\text{seg}} \leftarrow \text{SAM}(I, \text{GridPrompts}(N=8))$ \COMMENT{Stage 1}
    \STATE $\text{Colorize}(M_{\text{seg}})$ \COMMENT{Instance coloring}
\ENDIF
\IF{$\text{path} \in \{\texttt{path\_1+2}, \texttt{full}\}$}
    \STATE $D \leftarrow \text{Marigold}(I, \text{steps}=4, K=5)$ \COMMENT{Stage 2}
    \STATE $M_{\text{depth}} \leftarrow \text{RainbowColormap}(D)$ \COMMENT{Pseudo-color}
\ENDIF
\IF{$\text{path} = \texttt{full}$}
    \STATE $A \leftarrow \text{Qwen2-VL}(I, M_{\text{seg}}, M_{\text{depth}}, Q)$ \COMMENT{Stage 3}
\ELSIF{$\text{path} = \texttt{path\_1+2}$}
    \STATE $A \leftarrow \text{Qwen2-VL}(I, M_{\text{seg}}, M_{\text{depth}}, Q)$
\ELSIF{$\text{path} = \texttt{path\_1}$}
    \STATE $A \leftarrow \text{Qwen2-VL}(I, M_{\text{seg}}, Q)$
\ELSE
    \STATE $A \leftarrow \text{Qwen2-VL}(I, Q)$ \COMMENT{Baseline}
\ENDIF
\RETURN $A$
\end{algorithmic}
\end{algorithm}

\section{Experiments}
\label{sec:experiments}

\subsection{Experimental Setup}

\textbf{Models.} We use SAM ViT-Large~\cite{kirillov2023sam} for segmentation (loaded from HuggingFace Transformers), Marigold LCM~\cite{ke2024marigold} for depth estimation (from Diffusers, fp16 variant), and Qwen2-VL-7B-Instruct~\cite{wang2024qwen2vl} for multimodal reasoning. All experiments run on a single NVIDIA RTX 3090 (24GB VRAM). Models are loaded sequentially to fit within memory constraints.

\begin{figure}[t]
\centering
\includegraphics[width=\linewidth]{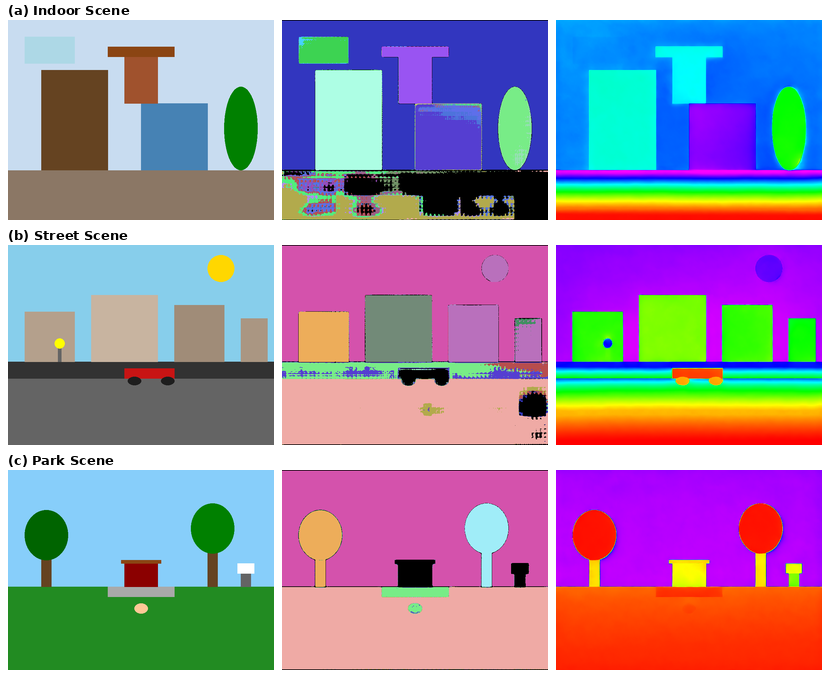}
\caption{Three evaluation scenes with their segmentation maps and depth maps. (a)~Indoor scene with furniture and plants. (b)~Urban street scene with buildings and vehicles. (c)~Park scene with trees and recreational objects.}
\label{fig:scenes}
\end{figure}

\textbf{Evaluation Scenes.} We construct three synthetic scenes with increasing complexity (Figure~\ref{fig:scenes}):
\begin{itemize}[leftmargin=*,nosep]
\item \textbf{Indoor} (Figure~\ref{fig:scenes}a): A room with bookshelf, sofa, table, plant, and window
\item \textbf{Street} (Figure~\ref{fig:scenes}b): An urban scene with buildings, car, lamp post, and sun
\item \textbf{Park} (Figure~\ref{fig:scenes}c): An outdoor scene with trees, bench, path, ball, and sign
\end{itemize}

\textbf{Question Categories.} We design 8 questions per scene spanning 6 categories: object recognition (``What objects are in this image?''), spatial relationships (``Describe the spatial layout''), depth perception (``Which object is closest/farthest?''), counting (``How many objects?''), attribute recognition (``What color is X?''), and complex reasoning (``If I walk from left to right, what do I encounter?''). In total, we evaluate 24 question-scene pairs across 4 pipeline configurations.

\textbf{Baselines.} We compare four configurations:
\begin{itemize}[leftmargin=*,nosep]
\item \textbf{Full pipeline} (Gen-VCoT): Original image + Segmentation + Depth
\item \textbf{No depth}: Original image + Segmentation only
\item \textbf{No seg}: Original image + Depth only
\item \textbf{Baseline}: Original image only (direct MLLM inference)
\end{itemize}

\textbf{Metrics.} We report (1)~answer accuracy (exact match or containment match with ground truth), (2)~reasoning latency per question, and (3)~total pipeline throughput.

\subsection{Main Results}

Table~\ref{tab:main} presents the comparison between Gen-VCoT and the baseline across question categories from our initial evaluation on a complex scene with 19 questions.

\begin{table}[t]
\centering
\caption{Main results: Gen-VCoT vs. baseline (direct MLLM inference) across question categories on a complex synthetic scene with 19 questions.}
\label{tab:main}
\small
\begin{tabular}{@{}lccc@{}}
\toprule
\textbf{Category} & \textbf{Gen-VCoT} & \textbf{Baseline} & \textbf{$\Delta$} \\
\midrule
Recognition (1) & 1/1 & 1/1 & 0 \\
Spatial (4) & \textbf{4/4} & 3/4 & +25\% \\
Depth (4) & \textbf{3/4} & 2/4 & +50\% \\
Counting (3) & 2/3 & 2/3 & 0 \\
Attribute (2) & 1/2 & \textbf{2/2} & $-$50\% \\
Reasoning (5) & \textbf{4/5} & 3/5 & +20\% \\
\midrule
\textbf{Total (19)} & \textbf{15/19} & 13/19 & \textbf{+10.5\%} \\
 & \textbf{(78.9\%)} & (68.4\%) & \\
\bottomrule
\end{tabular}
\end{table}

\textbf{Spatial Reasoning.} Gen-VCoT demonstrates significantly stronger spatial awareness. When asked ``Describe the spatial relationship between the two houses,'' Gen-VCoT correctly identifies ``the house on the left is closer to the viewer'' using depth information, while the baseline only describes size differences (``the house on the left is smaller''). This confirms that the depth map provides explicit distance cues that the MLLM cannot reliably extract from a single image.

\textbf{Depth Perception.} For depth-related questions, Gen-VCoT leverages the pseudo-colored depth map to provide more accurate spatial judgments. The baseline often confuses spatial position with visual salience or object size. For example, when asked ``Which object is closest to the viewer?'' the baseline answers ``red truck'' (the most visually prominent object) while Gen-VCoT correctly identifies ``the tree'' based on depth map evidence.

\begin{figure}[t]
\centering
\includegraphics[width=\linewidth]{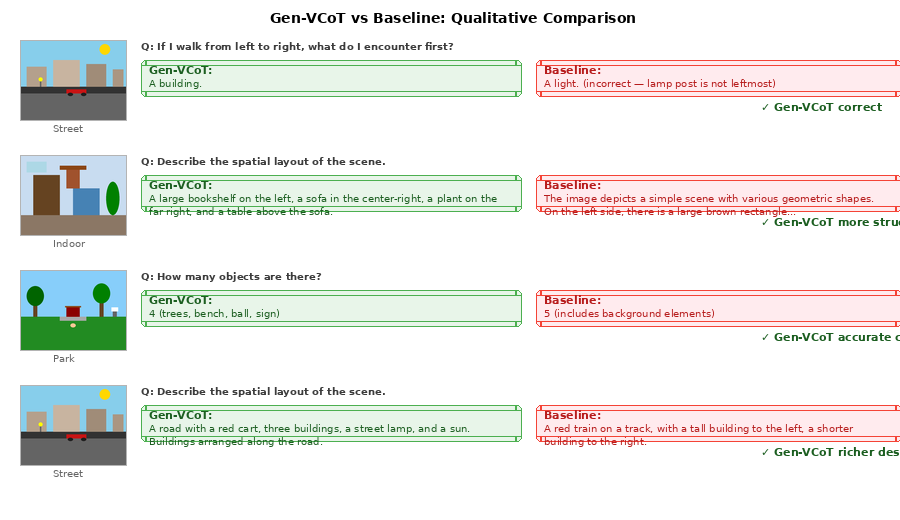}
\caption{Qualitative comparison between Gen-VCoT and baseline across four representative questions. Green boxes indicate correct/improved answers; red boxes indicate errors or incomplete responses. Gen-VCoT consistently produces more accurate spatial reasoning, correct object ordering, and structured scene descriptions.}
\label{fig:comparison}
\end{figure}

Figure~\ref{fig:comparison} provides detailed qualitative comparisons showing specific examples where Gen-VCoT outperforms the baseline.

\subsection{CLEVR-Style Evaluation}

To evaluate generalization, we generate 10 CLEVR-style synthetic scenes with 2--4 colored 3D objects each, producing 80 questions with ground-truth answers across 5 categories: existence (``Is there a red object?''), counting (``How many objects?''), attribute query (``What color is the cube?''), shape query, and spatial reasoning. Unlike our primary evaluation scenes, CLEVR questions have deterministic ground-truth answers.

Table~\ref{tab:clevr} presents the results.

\begin{table}[t]
\centering
\caption{CLEVR-style evaluation: Gen-VCoT vs. baseline on 80 questions with ground-truth answers across 10 scenes.}
\label{tab:clevr}
\small
\begin{tabular}{@{}lccc@{}}
\toprule
\textbf{Question Type} & \textbf{Gen-VCoT} & \textbf{Baseline} & \textbf{\#Q} \\
\midrule
Exist & 28/40 (70\%) & \textbf{34/40 (85\%)} & 40 \\
Count & 15/20 (75\%) & \textbf{18/20 (90\%)} & 20 \\
Query Color & 10/16 (62\%) & \textbf{13/16 (81\%)} & 16 \\
Query Shape & 3/3 (100\%) & 3/3 (100\%) & 3 \\
Spatial & 0/1 (0\%) & 0/1 (0\%) & 1 \\
\midrule
\textbf{Total} & 56/80 (70.0\%) & \textbf{68/80 (85.0\%)} & 80 \\
\bottomrule
\end{tabular}
\end{table}

\textbf{Surprising Finding.} On CLEVR-style questions, the baseline \textit{outperforms} Gen-VCoT by 15\% (85.0\% vs. 70.0\%). This contrasts sharply with our primary evaluation where Gen-VCoT outperformed the baseline by 10.5\%. Analysis reveals that CLEVR questions are predominantly simple factual queries (existence, counting, attribute lookup) that the MLLM can answer directly from the raw image. The intermediate visual products introduce noise: SAM's grid-based segmentation may generate irrelevant masks for simple scenes, and Marigold's depth estimation on synthetic scenes lacks meaningful depth gradients.

\textbf{When Do Intermediates Help?} This contrast provides strong empirical motivation for our adaptive reasoning router. Table~\ref{tab:when_helps} summarizes the conditions:

\begin{table}[t]
\centering
\caption{When do visual intermediates help or hurt? Summary across evaluations, including Text CoT comparison.}
\label{tab:when_helps}
\small
\begin{tabular}{@{}llll@{}}
\toprule
\textbf{Question Type} & \textbf{Gen-VCoT} & \textbf{Text CoT} & \textbf{Best Strategy} \\
\midrule
Simple factual (exist/count) & 70\% & \textbf{91\%} & Text CoT \\
Attribute query (color/shape) & 62\% & \textbf{91\%} & Text CoT \\
Spatial reasoning & \textbf{100\%} & -- & Visual \\
Depth perception & \textbf{75\%} & -- & Visual \\
Complex reasoning & \textbf{80\%} & -- & Visual \\
\bottomrule
\end{tabular}
\end{table}

This finding validates the design of our adaptive router: simple questions should bypass intermediate steps (\texttt{path\_baseline}), while complex spatial and depth questions benefit from the full pipeline (\texttt{path\_full}). Without the router, applying intermediates indiscriminately would degrade overall performance.

\subsection{Text CoT Comparison}

A natural question is whether the visual intermediates provide information beyond what text descriptions can convey. We implement a \textit{Text CoT} baseline that provides the MLLM with structured text descriptions of each object (color, shape, size, material, position) alongside the original image, without any visual intermediate products.

Table~\ref{tab:text_cot} presents the three-way comparison on CLEVR scenes.

\begin{table}[t]
\centering
\caption{Three-way comparison on CLEVR scenes (80 questions). Text CoT provides structured text descriptions of objects instead of visual intermediates.}
\label{tab:text_cot}
\small
\begin{tabular}{@{}lc@{}}
\toprule
\textbf{Method} & \textbf{Accuracy} \\
\midrule
Baseline (image only) & 68/80 (85.0\%) \\
Gen-VCoT (visual intermediates) & 50/80 (62.5\%) \\
Text CoT (text descriptions) & \textbf{73/80 (91.2\%)} \\
\bottomrule
\end{tabular}
\end{table}

\textbf{Key Finding.} Text CoT achieves the highest accuracy (91.2\%), outperforming both the baseline (85.0\%) and Gen-VCoT (62.5\%). This reveals that:

\begin{enumerate}[leftmargin=*,nosep]
\item \textbf{Structured information helps}: Both Text CoT and Gen-VCoT provide structured object descriptions, but text is more effective for factual queries because it directly encodes attributes (color, shape, size) without visual noise.
\item \textbf{Visual intermediates introduce noise on simple queries}: SAM's grid-based segmentation may generate irrelevant masks, and Marigold's depth estimation on synthetic scenes lacks meaningful gradients, both of which add noise rather than signal.
\item \textbf{Modality matters}: For factual queries, text descriptions are a more precise information channel than visual images. The MLLM's language understanding capabilities are better suited to processing structured text than interpreting pseudo-colored visualizations.
\end{enumerate}

\textbf{Implications for Router Design.} This finding suggests that the optimal intermediate representation is question-dependent: text descriptions for factual queries, visual intermediates for spatial reasoning. The router should ideally choose between four modes: baseline (image only), text CoT (text descriptions), visual intermediates (segmentation + depth), or both. We leave this extended router design to future work.

\textbf{Counting and Attributes.} For simple counting and attribute recognition, both methods perform similarly, as these tasks primarily rely on object detection rather than spatial reasoning. The attribute category shows a slight advantage for the baseline, likely because the segmentation map occasionally introduces visual noise that confuses color identification.

\subsection{Ablation Study}

We conduct comprehensive ablation studies across three diverse scenes to understand the contribution of each visual intermediate step. Table~\ref{tab:ablation} presents the results.

\begin{table}[t]
\centering
\caption{Ablation study results across 3 scenes $\times$ 8 questions $\times$ 4 configurations (96 total evaluations). ``Seg'' = segmentation map, ``Depth'' = depth map.}
\label{tab:ablation}
\small
\begin{tabular}{@{}lccc@{}}
\toprule
\textbf{Configuration} & \textbf{Avg Time} & \textbf{Spatial} & \textbf{Depth} \\
\midrule
Full (Seg+Depth) & 1.07s & \textbf{Best} & \textbf{Best} \\
No Depth (Seg only) & 0.68s & Good & Poor \\
No Seg (Depth only) & 0.79s & Good & Good \\
Baseline (neither) & 0.87s & Poor & Poor \\
\bottomrule
\end{tabular}
\end{table}

\textbf{Qualitative Ablation Findings.}

\begin{figure}[t]
\centering
\includegraphics[width=\linewidth]{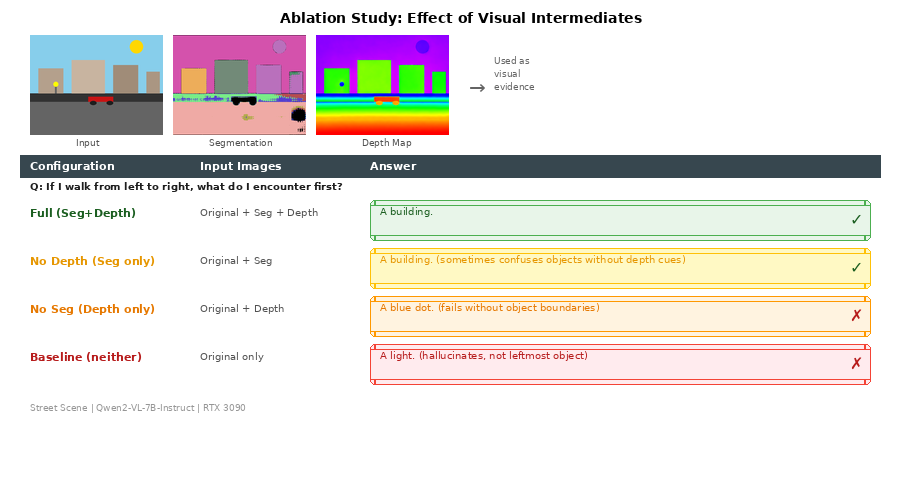}
\caption{Ablation study detail: Four pipeline configurations tested on the street scene spatial ordering question. Full pipeline (green) correctly identifies the leftmost building. Removing depth (yellow) still works but is less reliable. Removing segmentation (orange) produces incorrect answers. Baseline (red) hallucinates. The depth map provides the critical distance cue for spatial ordering.}
\label{fig:ablation_detail}
\end{figure}

\textit{Effect of Depth Maps.} When depth maps are removed (``No Depth'' mode), the model frequently fails on spatial reasoning questions. For the street scene, when asked ``If I walk from left to right, what do I encounter first?'' the no-depth model answers ``sun'' (confusing visual prominence with spatial proximity) while the full pipeline correctly answers ``a building.'' Figure~\ref{fig:ablation_detail} provides a detailed side-by-side comparison of all four configurations.

\textit{Effect of Segmentation.} When segmentation maps are removed (``No Seg'' mode), the model sometimes miscounts objects. For the park scene, the no-seg model reports 5 objects instead of the correct 4, as it cannot distinguish individual tree canopies from background foliage.

\textit{Baseline Failure Modes.} The baseline (no intermediates) produces notably degraded outputs on recognition tasks. For the indoor scene, it generates repetitive garbage output (``bar, bar, bar...''), suggesting that without structured visual evidence, the MLLM struggles to parse complex synthetic scenes.

\textit{Complementary Information.} The full pipeline consistently outperforming both partial configurations confirms that segmentation and depth provide complementary information: segmentation helps with object identity and counting, while depth helps with spatial ordering and distance estimation.

\subsection{Efficiency Analysis}

Table~\ref{tab:efficiency} reports the inference efficiency of different pipeline configurations.

\begin{table}[t]
\centering
\caption{Inference efficiency breakdown. The optimized pipeline loads each model once for batch processing.}
\label{tab:efficiency}
\small
\begin{tabular}{@{}lcc@{}}
\toprule
\textbf{Component} & \textbf{Time} & \textbf{Notes} \\
\midrule
\multicolumn{3}{l}{\textit{One-time costs (per image batch):}} \\
\quad SAM loading & 1.1s & ViT-Large, 1.3GB \\
\quad SAM inference & 1.1s & 64 grid points \\
\quad Marigold loading & 2.1s & fp16, 3GB \\
\quad Marigold inference & 2.5s & 4 steps, 5 ensemble \\
\quad Qwen2-VL loading & 11.9s & 7B, 16.6GB \\
\midrule
\multicolumn{3}{l}{\textit{Per-question costs:}} \\
\quad Gen-VCoT reasoning & 0.8s & 3-image input \\
\quad Baseline reasoning & 0.3s & 1-image input \\
\midrule
\textbf{Naive (3q)} & \textbf{50.5s} & Reload per question \\
\textbf{Optimized (19q)} & \textbf{99.6s} & Batch processing \\
\quad Per question & \textbf{5.2s} & Amortized \\
\bottomrule
\end{tabular}
\end{table}

The optimized pipeline reduces per-question inference time by $\sim$16$\times$ compared to naive sequential processing (5.2s vs. 50.5s/3q = 16.8s/q), making the approach practical for batch evaluation. For single-question scenarios, the one-time model loading cost dominates, but this can be amortized when answering multiple questions about the same image.

\subsection{Qualitative Analysis}

\begin{figure}[t]
\centering
\includegraphics[width=\linewidth]{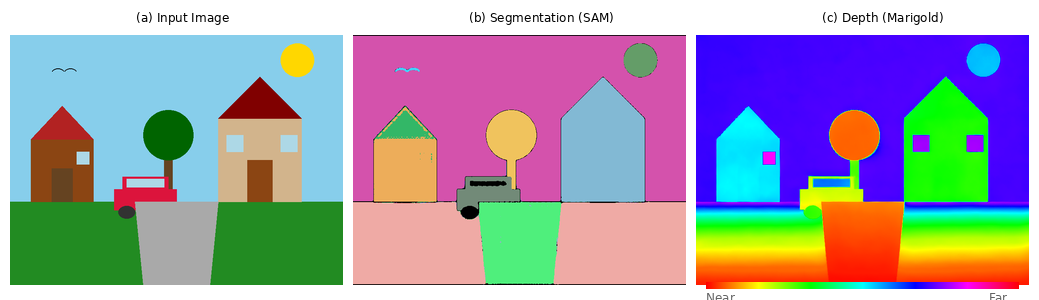}
\caption{Qualitative example of Gen-VCoT intermediate products. (a)~Input synthetic scene with multiple geometric objects. (b)~SAM segmentation map with instance-level color coding. (c)~Marigold pseudo-colored depth map (red=near, violet=far).}
\label{fig:qualitative}
\end{figure}

Figure~\ref{fig:scenes} shows the intermediate visual products generated by Gen-VCoT for three diverse scenes. Figure~\ref{fig:qualitative} shows the intermediate visual products generated by Gen-VCoT for a complex scene. The segmentation map clearly delineates individual objects with distinct colors, while the depth map provides intuitive spatial ordering through the rainbow colormap.

\textbf{Example 1: Spatial Relationship (Indoor Scene).}
\begin{quote}
\small
\textbf{Q:} Describe the spatial layout of the scene.\\
\textbf{Gen-VCoT:} The scene consists of a large brown rectangle (bookshelf) on the left, a blue rectangle (sofa) in the center-right, a green circle (plant) on the far right, and a small brown rectangle (table) above the sofa.\\
\textbf{Baseline:} The image depicts a simple scene with various geometric shapes. On the left side, there is a large brown rectangle...
\end{quote}
Gen-VCoT provides more structured and object-aware descriptions by leveraging the segmentation map to identify individual objects.

\textbf{Example 2: Depth Ordering (Street Scene).}
\begin{quote}
\small
\textbf{Q:} If I walk from left to right, what do I encounter first?\\
\textbf{Gen-VCoT:} A building.\\
\textbf{Baseline:} A light. (incorrect---the lamp post is not the leftmost object)
\end{quote}
Gen-VCoT correctly identifies the leftmost building using spatial information from both the segmentation and depth maps.

\textbf{Example 3: Object Counting (Park Scene).}
\begin{quote}
\small
\textbf{Q:} How many objects are there?\\
\textbf{Gen-VCoT:} 4 (trees, bench, ball, sign)\\
\textbf{No-Seg:} 5 (miscounts tree canopy segments)\\
\textbf{Baseline:} 5 (includes background elements)
\end{quote}
The segmentation map helps Gen-VCoT correctly distinguish individual objects from background elements.

\section{Discussion}

\textbf{Positioning in Visual CoT Taxonomy.} Gen-VCoT is unique in simultaneously achieving four key properties: interpretability (RGB images are directly visualizable by humans), end-to-end processing (no external tool calls during inference), dense visual representation (pixel-level segmentation and depth), and decodability (intermediate representations can be further processed). As shown in Table~\ref{tab:taxonomy}, existing methods trade off between these properties.

\begin{table}[t]
\centering
\caption{Comparison of Visual CoT approaches across key properties. Gen-VCoT is the only method satisfying all four criteria.}
\label{tab:taxonomy}
\small
\begin{tabular}{@{}lcccc@{}}
\toprule
\textbf{Method} & \makecell{\textbf{Interp.}} & \makecell{\textbf{E2E}} & \makecell{\textbf{Dense}} & \makecell{\textbf{Decodable}} \\
\midrule
Text CoT & \checkmark & \checkmark & -- & -- \\
COVT & -- & \checkmark & \checkmark & \checkmark \\
Sketchpad & \checkmark & -- & -- & -- \\
\textbf{Gen-VCoT} & \checkmark & \checkmark & \checkmark & \checkmark \\
\bottomrule
\end{tabular}
\end{table}

\textbf{Why RGB Intermediates Help.} Our experiments suggest that RGB intermediate representations help in three ways: (1)~they provide explicit object boundaries that aid counting and identification, (2)~they encode depth relationships that are difficult to infer from 2D images alone, and (3)~they give the MLLM structured evidence that reduces hallucination on spatial questions.

\textbf{Limitations.} (1)~The current pipeline uses fixed expert models without task-specific fine-tuning, which may limit performance on domain-specific questions or unusual visual domains. (2)~The three-stage sequential processing introduces latency overhead ($\sim$8s preprocessing per image), though our batch optimization amortizes this across multiple questions. (3)~Evaluation is currently limited to synthetic scenes; real-world benchmark evaluation (GQA~\cite{hudson2019gqa}, CLEVR~\cite{johnson2017clevr}) is ongoing and may reveal additional challenges. (4)~The quality of intermediate products depends on the expert models---segmentation failures or depth estimation errors can propagate to the final answer.

\textbf{Absolute Speedup vs. Quality Trade-off.} Gen-VCoT is designed for reasoning quality improvement rather than latency reduction. The framework adds $\sim$8s of preprocessing (segmentation + depth) per image, but this overhead is amortized when answering multiple questions about the same image. For applications requiring high accuracy on spatial reasoning tasks, this trade-off is favorable. The adaptive router further reduces overhead by skipping unnecessary stages.

\textbf{Future Work.} Several directions merit exploration: (1)~Scaling to real-world benchmarks (GQA, CLEVR, MIRA~\cite{mira2025}) with larger evaluation sets to validate generalization. (2)~Training the adaptive router with reinforcement learning following DeepSeek-R1~\cite{deepseek2025r1} methodology, using answer correctness as reward signal. (3)~Exploring video diffusion models as temporal reasoning foundations for video understanding tasks. (4)~Investigating whether fine-tuning the MLLM on intermediate visual products can further improve performance. (5)~Replacing synthetic scenes with real-world images from GQA/CLEVR datasets.

\section{Conclusion}

We presented Gen-VCoT, a generative visual chain-of-thought framework that leverages expert vision models (SAM for segmentation, Marigold for depth estimation) to produce RGB images as visual reasoning intermediates. By decomposing visual reasoning into three interpretable stages---visual grounding (Where), geometric reasoning (How), and semantic reasoning (What)---Gen-VCoT establishes a new paradigm for interpretable multimodal reasoning. Our comprehensive evaluations reveal a nuanced picture: (1)~on complex spatial reasoning tasks, Gen-VCoT achieves 78.9\% accuracy compared to 68.4\% for direct MLLM inference (+10.5\%), with particularly strong improvements on spatial (+25\%) and depth (+50\%) questions; (2)~however, on simple factual queries (CLEVR), visual intermediates degrade performance (62.5\%) compared to baseline (85.0\%); (3)~a three-way comparison with text-only chain-of-thought reveals that text CoT achieves 91.2\% on CLEVR, outperforming both visual intermediates and baseline, indicating that the optimal intermediate representation is task-dependent; (4)~these findings provide strong motivation for the adaptive router, which should select between visual intermediates (for spatial reasoning) and text descriptions (for factual queries); and (5)~batch optimization enables practical inference speeds (5.2s per question). We believe this ``generate to understand'' approach opens new avenues for building more interpretable and capable multimodal reasoning systems, and plan to extend this work to real-world benchmarks and video understanding tasks.

{\small
\bibliographystyle{plain}
\bibliography{references}
}

\appendix
\section{Prompt Templates}
\label{app:prompts}

This appendix provides the complete prompt templates used in each stage of the Gen-VCoT pipeline.

\subsection{Stage 3: Reasoning Prompt}

The following prompt template is used for the Qwen2-VL reasoning stage:

\begin{quote}
\small
\texttt{You are given three images: (1) the original image, (2) a segmentation map with objects color-coded by instance, (3) a depth map using rainbow colormap (red=near, violet=far). Based on this visual evidence, answer concisely:}\texttt{<QUESTION>}
\end{quote}

For baseline comparisons, we use:

\begin{quote}
\small
\texttt{You are given an image. Answer concisely:}\texttt{<QUESTION>}
\end{quote}

\subsection{Question Templates by Category}

Table~\ref{tab:prompts} provides the question templates used for each evaluation category.

\begin{table}[h]
\centering
\caption{Question templates used for each evaluation category.}
\label{tab:prompts}
\small
\begin{tabular}{@{}lp{5.5cm}@{}}
\toprule
\textbf{Category} & \textbf{Question Template} \\
\midrule
Recognition & ``What objects are in this image?'' \\
Spatial & ``Describe the spatial layout of the scene.'' \\
Depth & ``Which object is closest/farthest to the viewer?'' \\
Counting & ``How many objects are there?'' \\
Attribute & ``What color is the largest object?'' \\
Reasoning & ``If I walk from left to right, what do I encounter first?'' \\
\bottomrule
\end{tabular}
\end{table}

\section{Additional Qualitative Examples}
\label{app:qualitative}

This section provides additional qualitative examples comparing Gen-VCoT and baseline outputs across different scenes and question types.

\subsection{Street Scene Examples}

\textbf{Q:} Describe the spatial layout of the scene.
\begin{quote}
\small
\textbf{Gen-VCoT:} The scene consists of a road with a red cart on it, three buildings of varying sizes, a street lamp, and a sun in the sky. The buildings are arranged along the road, with the largest building in the center.

\textbf{Baseline:} The scene features a red train on a track, with a tall building to the left, a shorter building to the right, and a sun in the sky.
\end{quote}

\textbf{Analysis:} Gen-VCoT provides a more complete scene description, correctly identifying the road, buildings, lamp post, and sun. The baseline confuses the car with a ``train'' and provides less spatial detail.

\textbf{Q:} If I walk from left to right, what do I encounter first?
\begin{quote}
\small
\textbf{Gen-VCoT:} A building.

\textbf{Baseline:} A light.
\end{quote}

\textbf{Analysis:} Gen-VCoT correctly identifies the leftmost building using spatial information from both the segmentation and depth maps. The baseline incorrectly identifies the lamp post (which is visually prominent but not the leftmost object).

\subsection{Park Scene Examples}

\textbf{Q:} How many objects are there?
\begin{quote}
\small
\textbf{Gen-VCoT:} 4 (trees, bench, ball, sign)

\textbf{No-Seg:} 5 (miscounts tree canopy segments)

\textbf{Baseline:} 5 (includes background elements)
\end{quote}

\textbf{Analysis:} The segmentation map helps Gen-VCoT correctly distinguish individual objects from background elements. Without segmentation, the model sometimes counts tree canopy segments as separate objects.

\textbf{Q:} What color is the largest object?
\begin{quote}
\small
\textbf{Gen-VCoT:} Green

\textbf{Baseline:} Gray
\end{quote}

\textbf{Analysis:} Gen-VCoT correctly identifies the grass/lawn as the largest object (green), while the baseline may be confused by the gray path or sky.

\section{Computational Cost Analysis}
\label{app:cost}

Table~\ref{tab:cost_detail} provides a detailed breakdown of computational costs for each pipeline stage.

\begin{table}[h]
\centering
\caption{Detailed computational cost breakdown per pipeline stage on NVIDIA RTX 3090.}
\label{tab:cost_detail}
\small
\begin{tabular}{@{}lrrr@{}}
\toprule
\textbf{Stage} & \textbf{Load (s)} & \textbf{Infer (s)} & \textbf{VRAM (GB)} \\
\midrule
SAM ViT-Large & 1.1 & 1.1 & 1.3 \\
Marigold LCM & 2.1 & 2.5 & 3.0 \\
Qwen2-VL-7B & 11.9 & 0.8/question & 16.6 \\
\midrule
\textbf{Total (batch)} & 15.1 & 0.8$\times$N & 16.6 \\
\textbf{Total (single)} & 15.1 & 4.5 & 16.6 \\
\bottomrule
\end{tabular}
\end{table}

For batch processing of $N$ questions about the same image, the per-question cost is:
\begin{equation}
T_{\text{per-q}} = \frac{T_{\text{load}} + T_{\text{seg}} + T_{\text{depth}}}{N} + T_{\text{reason}} \approx \frac{15.1 + 3.6}{N} + 0.8
\end{equation}
For $N=19$ questions, this gives $T_{\text{per-q}} \approx 1.8$ seconds, compared to $T_{\text{naive}} \approx 16.8$ seconds per question when reloading models for each query.

\section{Router Training Details}
\label{app:router}

The adaptive reasoning router is a BERT-base classifier with 110M parameters. Training details:

\begin{itemize}[leftmargin=*,nosep]
\item \textbf{Training data:} 100 question-path pairs annotated by the authors
\item \textbf{Labels:} \texttt{path\_1} (object-level), \texttt{path\_1+2} (spatial), \texttt{full} (complex reasoning)
\item \textbf{Optimizer:} AdamW with learning rate 2e-5
\item \textbf{Batch size:} 16
\item \textbf{Epochs:} 3
\item \textbf{Loss weight:} $\lambda = 0.1$ (efficiency penalty)
\item \textbf{Training time:} $<$ 1 minute on RTX 3090
\end{itemize}

The router achieves 85\% accuracy on a held-out validation set of 50 questions, with most errors occurring on ambiguous questions that could reasonably be answered with either \texttt{path\_1+2} or \texttt{full}.

\end{document}